\documentclass{bmvc2k}

\title{Automatic Semantic Content Removal \\by Learning to Neglect}

\addauthor{Siyang Qin}{siqin@soe.ucsc.edu}{1}
\addauthor{Jiahui Wei}{jwei19@ucsc.edu}{1}
\addauthor{Roberto Manduchi}{manduchi@soe.ucsc.edu}{1}

\addinstitution{
 University of California, Santa Cruz\\
 Santa Cruz, CA, USA
}

\runninghead{Qin, Wei, Manduchi}{Automatic Semantic Content Removal}

\usepackage{graphicx}
\usepackage{amssymb}
\usepackage{amsmath}
\usepackage{epsfig}

\def\eg{\emph{e.g}\bmvaOneDot}

\def\etal{\emph{et al}\bmvaOneDot}

\begin{document}

\maketitle

\begin{abstract}
We introduce a new system for automatic image content removal and inpainting. Unlike traditional inpainting algorithms, which require advance knowledge of the region to be filled in, our system automatically detects the area to be removed and infilled. Region segmentation and inpainting are performed jointly in a single pass. In this way, potential segmentation errors are more naturally alleviated by the inpainting module. The system is implemented as an encoder-decoder architecture, with two decoder branches, one tasked with segmentation of the foreground region, the other with inpainting. The encoder and the two decoder branches are linked via {\em neglect nodes}, which guide the inpainting process in selecting which areas need reconstruction. The whole model is trained using a conditional GAN strategy. Comparative experiments show that our algorithm outperforms state-of-the-art inpainting techniques (which, unlike our system, do not segment the input image and thus must be aided by an external segmentation module.)
\end{abstract}

\section{Introduction}
\label{sec:intro}
 Automatic  removal of specific content in an image is a task of practical interest, as well as of intellectual appeal.
There are many situations in which a part of an image needs to be erased and replaced. This may include text (whether present in the scene or overimposed on the image), which may have to be removed, for example to protect personal information; people, such as undesired passersby in the scene; or other objects that, for any reasons, one may want to wipe off the picture. While these operations are typically performed manually by skilled Photoshop editors, substantial cost reduction could be achieved by automatizing the workflow. Automatic content removal, though, is difficult, and an as yet unsolved problem. Removing content and inpainting the image in a way that it looks ``natural'' entails the ability to capture, represent, and synthesize  high-level (``semantic'') image content. This is particularly true of large image areas infilling, an operation that  only recently has been accomplished with some success~\cite{pathak2016context,yeh2017semantic,yang2017high,yu2018generative}.  

Image inpainting algorithms described in the literature normally require that a binary ``mask'' indicating the location of the area to be synthesized be provided, typically via manual input. In contrast, an automatic content removal system must be able to accomplish two  tasks.  First, the pattern or object of interest must be segmented out, creating a binary mask; then, the  image content within the mask must be synthesized. The work described in this paper is born from the realization that, for optimal results, these two tasks (segmentation and inpainting) should {\em not} be carried out independently.
In fact, only in few specific situations can the portion of the image to be removed be represented by a binary mask. Edge smoothing effects are almost always present, either due to the camera's point spreading function, or due to  blending, if the pattern (\eg text) is overimposed on the image. 
Although one could potentially recover an alpha mask for the foreground content to be removed, 
we believe that a more appropriate strategy is to {\em simultaneously} detect the foreground {\em and} synthesize the background image. By doing so, we do not need to resort to hand-made tricks, such as expanding the binary mask to account for inaccurate localization.

Our algorithm for content removal and inpainting relies on conditional generative adversarial networks (cGANs) \cite{isola2017image}, which have become the tool of choice for image synthesis. Our network architecture is based on an encoder-decoder scheme with skip layers (see Fig. \ref{fig:gan}). Rather than a single decoder branch, however, our network contains two parallel and interconnected branches: one ({\em dec-seg}) designed to extract the foreground area to be removed; the other  ({\em dec-fill}) tasked with synthesizing the missing background. The {\em dec-seg} branch interacts with the {\em dec-fill} branch via multiple {\em neglect nodes} (see Fig. \ref{fig:neglect}). The concept of neglect nodes is germane (but in reverse) to that of  mixing nodes normally found in { attention networks} \cite{xu2015show,vaswani2017attention}. Mixing nodes highlight a portion of the image that needs to be attended to, or,  in our case, neglected. Neglect nodes appear in all layers of the architecture; they ensure that the {\em dec-fill} branch is aware of which portions of the image are to be synthesized, without ever committing to a binary mask. 

A remarkable feature of the proposed system is that the multiple components of the network (encoder and two decoder branches, along with the neglect nodes) are all trained at the same time. Training seeks to minimize a global cost function that includes a conditional GAN component, as well as $L_1$ distance components for both foreground segmentation and background image. This optimization requires ground--truth availability of  {\em foreground} segmentation (the component to be removed), {\em background}  images (the original image without the foreground), and {\em composite} images (foreground over background).
By jointly optimizing the multiple network components (rather than, say, optimizing for the {\em dec-seg} independently on foreground segmentation, then using it to condition optimization of {\em dec-fill} via the neglect nodes), we are able to accurately reconstruct the  background inpainted image. The algorithm also produces the foreground segmentation as a byproduct. We should emphasize that this foreground mask is {\em not} used by the {\em dec-fill} synthesizing layer, which only communicates with the {\em dec-seg} layer via the neglect nodes.

To summarize, this paper has two main contributions. First, we present the first (to the best of our knowledge) truly automatic semantic content removal system with  promising results on  realistic images. The proposed algorithm is able to recover high-quality background without any knowledge of the foreground segmentation mask. Unlike most previous GAN--based inpainting methods that assume a rectangular  foreground region to be removed \cite{pathak2016context,yeh2017semantic,yang2017high}, our system produces good result with any foreground shape, even when it extends to the image boundary. Second, we introduce a novel encoder-decoder network structure with two parallel and interconnected branches ({\em dec-seg} and {\em dec-fill}),  linked at multiple levels by mixing (neglect) nodes that determine which information from the encoder should be used for synthesis, and which should be neglected.  Foreground region segmentation  and  background inpainting is produced in one single forward pass.

\section{Related Work}
\label{sec::rw}
Semantic image content removal comprises two different tasks:  segmentation of the foreground region (which in some contexts represents a ``corrupted'' image region to be removed), and synthesis of the missing background after foreground removal. We briefly review the literature for these two operations in the following.

Pixel-level semantic image segmentation has received considerable attention over the past few years. Most recently published techniques are based on fully convolutional networks (FCN) \cite{long2015fully}, possibly combined with fully connected CRF  \cite{krahenbuhl2011efficient,zheng2015conditional,chen2016deeplab,qin2018stroke}.
The general architecture of a FCN includes a sequence of convolution and downsampling layers ({\em encoder}), which extract multi--scale features, followed by a sequence of deconvolution layers ({\em decoder}), which produce a high--resolution segmentation (or ``prediction''). Skip layers are often added, providing shortcut links from an encoder layer to its corresponding decoder layer.  The role of  skip layers is to provide well-localized information to a decoder layer, in addition to the semantic-rich but poorly resolved information from the prior decoder layer. 
In this way, skip layers enable good pixel-level localization while facilitating gradient flow during training. 
Similar architectures have been used in various applications such as text segmentation \cite{qin2017cascaded,zhang2016multi,qin2018stroke,yao2016scene}, edge detection \cite{xie2015holistically}, face segmentation \cite{qin2017icme}, and scene parsing \cite{zhao2017pyramid}. Although these algorithms could be used for the foreground segmentation component of a content removal system, unavoidable inaccuracies are liable to 
dramatically decrease the quality of the recovered background region.

Image inpainting \cite{bertalmio2000image} has a long history. The goal of inpainting is to fill in a missing region with realistic image content. A variety of inpainting methods have been proposed, including those based on prior image statistics \cite{roth2005fields,zoran2011learning}, and those based on CNNs \cite{ren2015shepard,zhang2017learning}. More recently, outstanding results have been demonstrated with the use of GANs to inpaint even large missing areas  \cite{yeh2017semantic, yang2017high, pathak2016context,yu2018generative}. While appropriate for certain domains (\eg face inpainting), methods in this category often suffer from serious limitations, including the requirement that the missing region have fixed size and shape.

All of the inpainting methods mentioned above assume that the corrupted region mask is known (typically as provided by the user).  This limits their scope of application, as in most cases, this mask is unavailable. This shortcoming is addressed by blind inpainting algorithms \cite{xie2012image, liu2017deep}, which do not need access to the foreground mask. However, prior blind inpainting work has been demonstrated only for very simple cases, with constant-valued foreground occupying  a small area of the image. 

\section{Proposed Algorithm}
\label{sec::method}
Our system for  automatic semantic content removal comprises an encoder-decoder network with two decoder branches, tasked with  predicting a segmentation mask ({\em dec-seg}) and a background image ({\em dec-fill}) in a single forward pass. Neglect nodes (an original feature of this architecture) link the two decoder branches and the encoder at various levels. The network is trained along with a discriminator network in an adversarial scheme, in order to foster realistic background image synthesis.


We  assume in this work that the foreground region to be removed occupies a large portion of the image (or, equivalently, that the image is cropped such that the foreground region takes most of the cropped region). In practice, this can be obtained using a standard object detector. Note that high accuracy of the (rectangular) detector is not required. In our experiments, the margin between the contour of the foreground region and the edges of the image was let to vary between 0  (foreground touching the image edge) to half the size of the foreground mask.


\subsection{Loss Function}
\label{sec::loss}
Following the terminology of GANs, the output $z_p$ of {\em dec-seg} and $y_p$ of {\em dec-fill} for an input image $x$ are taken to represent the output of a generator $G(x)$. The generator is trained with a dual task: ensuring that $z_p$ and $y_p$ are similar to the ground--truth ($z_g$ and $y_g$), and deceiving the discriminator $D(x,y)$, which is concurrently trained to  separate $y_p$ from $y_g$ given $x$. The  cost function $L_G$ for the generator combines the conditional GAN loss with a linear combination of the $L_1$ distances between  prediction and  ground--truth for segmentation and inpainted background:


\begin{equation}
\label{loss::G}
L_G = \mathbb{E}[\log(1-D(x,y_p))]+\lambda_f\mathbb{E}[||y_g-y_p||_1]+\lambda_s\mathbb{E}[||z_g-z_p||_1]
\end{equation}

\noindent The discriminator $D$ is trained to minimize the following discriminator loss $L_D$:

\begin{equation}
\label{loss::D}
L_D = -(\mathbb{E}[\log D(x,y_g)]+\mathbb{E}[\log(1-D(x,y_p))])
\end{equation}

\begin{figure*}
\begin{center}
\includegraphics[width=\linewidth]{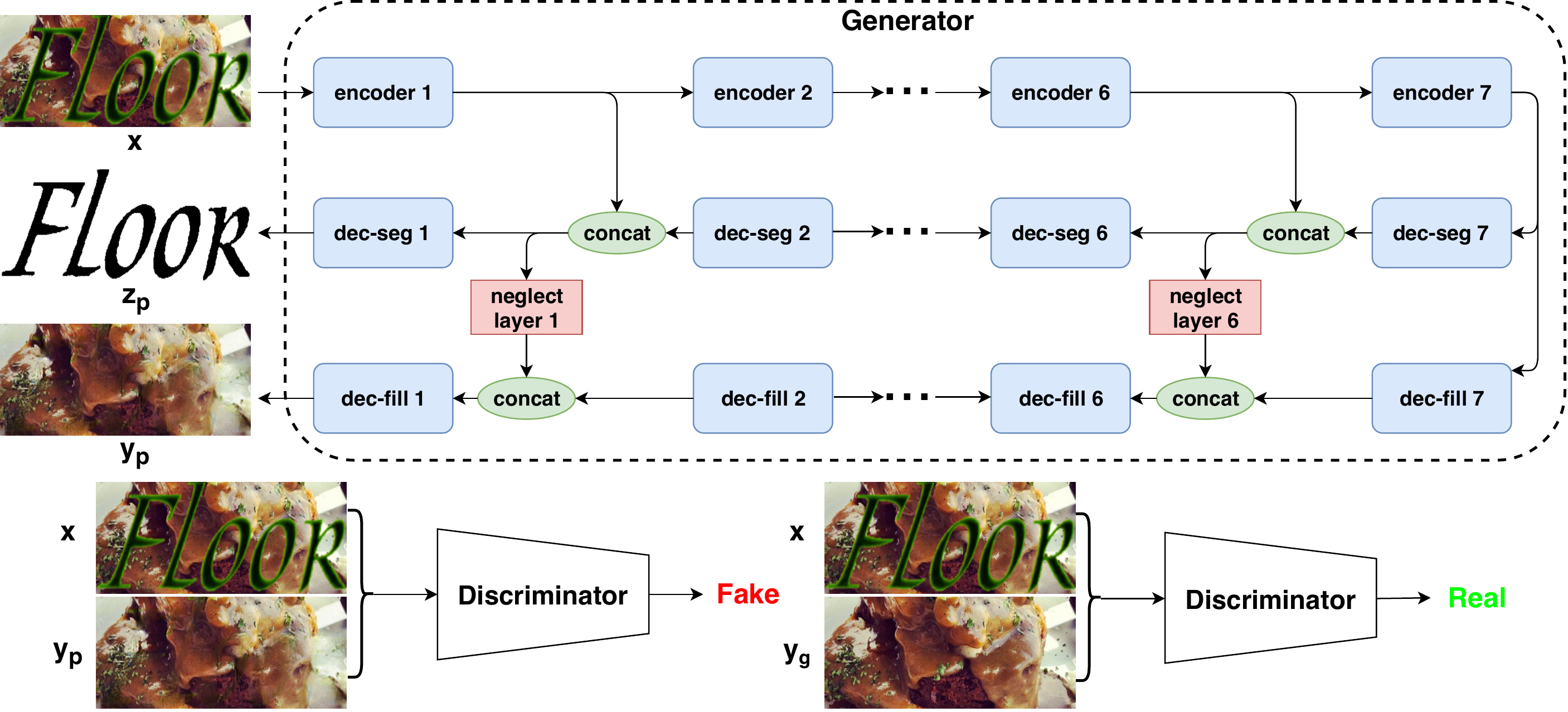}
\end{center}
\caption{System architecture.}
\label{fig:gan}
\end{figure*}

\subsection{Network Architecture}\label{sec:NA}
\textbf{Generator:} Segmentation and background infilling is generated in a single pass by an encoder--decoder network architecture, with multiple encoder layers  generating  multi-scale features at decreasing resolution, and two parallel decoder branches ({\em dec-seg} and {\em dec-fill}) producing high--resolution output starting from low--resolution features and higher--resolution  data from skip layers. 
Each encoder stage consists of a convolution layer with kernel of size $4\times 4$, stride $2$, followed by instance normalization \cite{ulyanov2016instance}  and ReLU. Each stage of  {\em dec-seg} contains a deconvolution (transpose convolution) layer (kernel of $4\times 4$, stride $2$), followed by  instance normalization and ReLU. {\em dec-fill} replaces each deconvolution layers with a nearest--neighbor upsampling layer followed by a convolution layer (kernel sized $3\times 3$, stride $1$). This strategy, originally proposed by Odena \etal \cite{odena2016deconvolution} to reduce checkerboard artifacts, was found to be very useful in our experiments (see Sec. \ref{sec:ablation}). The total number of convolution kernels at the $i$-th encoder layer is $\min(2^{i-1}\times 64, 512)$. The number of deconvolution kernels at the $i$-th {\em dec-seg} layer or convolution kernels at the $i$-th {\em dec-fill} layer are the same as the number of kernels at the $(i-1)$-th encoder layer.
The output of the first {\em dec-seg} layer and {\em dec-fill} layer has one channel (foreground segmentation) and three channels (recovered background image) respectively.
All ReLU layers in the encoder are leaky, with slope of $0.2$. In the {\em dec-seg} branch, standard skip layers are added. More precisely,  following the layer indexing in Fig. \ref{fig:gan}, the input of the $i$-th layer of {\em dec-seg} is a concatenation of the output of the  $(i+1)$-th  layer in the same branch and of the output of the $i$-th encoder layer (except for the 7-th decoder layer, which only receives input from the 7-th encoder layer.) As mentioned earlier, skip layers ensure good segmentation localization.

\begin{figure}
\begin{tabular}{cc}
\includegraphics[width=0.17\linewidth]{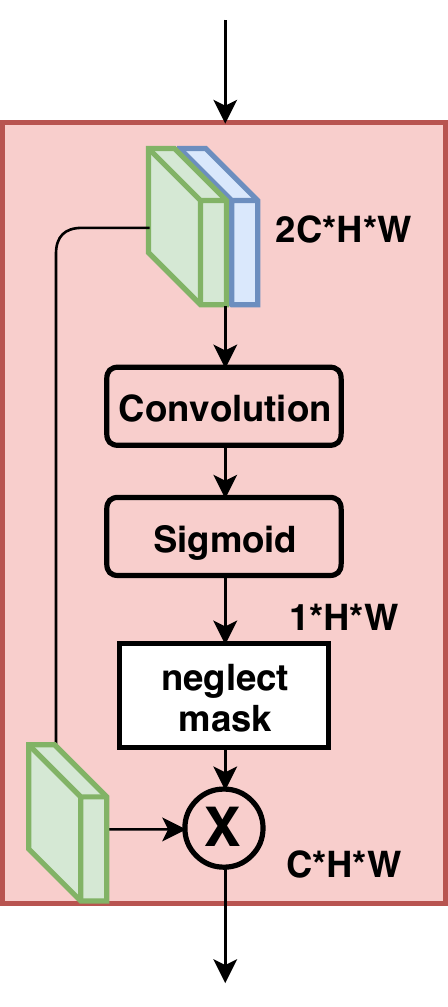}&
\includegraphics[width=0.78\linewidth]{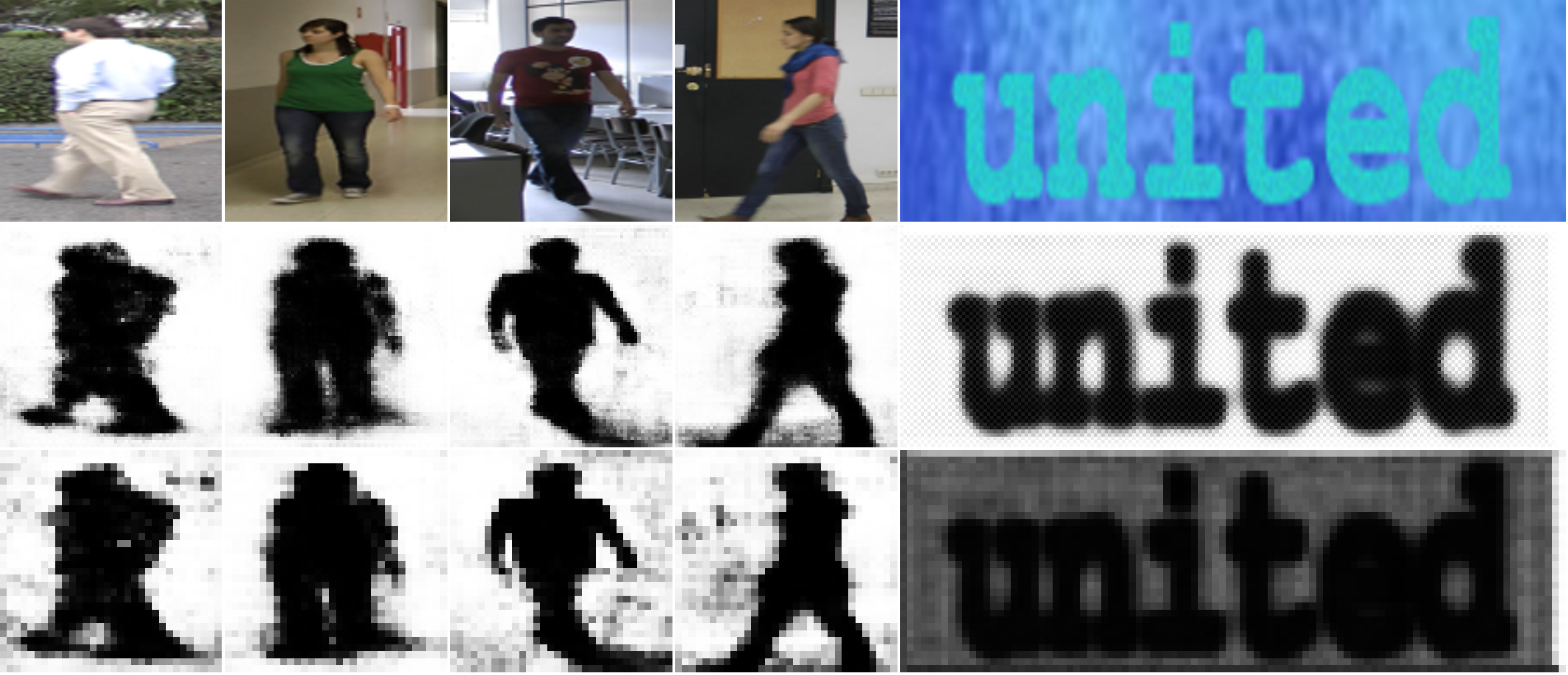} \\
(a)&(b)
\end{tabular}
\caption{(a) Neglect node architecture. Green blocks: encoder output. Blue block: output of previous {\em dec-seg} layer. (b) Top row: input images. Second and third row: associated neglect masks generated by the neglect node in layer 1 and 2, respectively.
}
\label{fig:neglect}
\end{figure}

The layers of {\em dec-fill} also receive information from equi-scale encoder layers, but this information is modulated by {\em neglect masks} generated by neglect nodes. 
 Specifically,  the $i$-th neglect node receives in input  data from the $i$-th encoder layer, concatenated with data  from the $(i+1)$-th {\em dec-seg} layer (note that this is the same as the input to the $i$-th {\em dec-seg} layer).  A $1\times 1$ convolution, followed by a sigmoid, produces a 
 neglect mask (an image with values between 0 and 1). The neglect mask modulates (by pixel multiplication) the content of the  $i$-th encoder layer, before it is concatenated with the output of the $(i+1)$-th {\em dec-fill}  layer and fed to the $i$-th {\em dec-fill}  layer. The process is shown in Fig.~\ref{fig:neglect}~(a).
  In practice, neglect nodes provide {\em dec-fill} with information about which areas of the image should be erased and infilled, while preserving content elsewhere. 
 Visual inspection of the neglect masks  shows that they faithfully identify the portion of the image to be removed at various scales (see \eg Fig.~\ref{fig:neglect}).
 


\vspace{2mm}
\noindent\textbf{Discriminator:} The input to the discriminator is the concatenation of the input image $x$ and of the predicted background $y_p$ or background ground--truth $y_g$. The structure of the discriminator is the same as the first 5 encoder layers of the generator, but its output undergoes a $1\times 1$ convolution layer followed by  a sigmoid function. In the case of $128\times 128$ input dimension, the output size is $4\times 4$, with values between 0 and 1, representing the decoder's confidence that the corresponding region is realistic. The average of these 16  values forms the final discriminator output.

\section{Experiments}
\subsection{Datasets}\label{sec:dataset}
Training our model requires, for each image, three pieces of information: the original background image (for {\em dec-fill}); the foreground region (mask) to be removed (for {\em dec-seg}); and the composite image. Given that this type of rich information is  not available in existing datasets, we built our own training and test sets. Specifically, we considered two different datasets for our experiments: one with synthetic text overimposed on regular images, and one with real images of pedestrians.

\vspace{2mm}
\noindent
\textbf{Text dataset:} Images in this dataset are generated by pasting text (generated synthetically) onto real background images. In this way, we have direct access to all necessary data (foreground, background, and composite image). 
Text images come from two resources: (1) the word synthesis engine of \cite{qin2018stroke}, which was used to generate 50K word images, along with the ground--truth associated segmentation masks; (2)  the ICDAR 2013 dataset \cite{icdarfocused}, which provides pixel-level text stroke labels, allowing us to extract 1850 real text regions.
Random geometry transformations and color jittering were used to augment the real text, obtaining 50K more word images. Given a sample from the 100K word image pool, a similarly sized background image patch was cropped at random positions from images randomly picked from the MS COCO dataset\cite{lin2014microsoft}. More specifically, the background images for  our training and validation set come from the training portion of the MS COCO dataset, while the background images for  our test set come from the testing portion of the MS COCO dataset. This ensures that the training and testing sets do not share background images. In total, the training, validation and testing portions of our synthetic text dataset contain 100K, 15K and  15K images, respectively.

\vspace{2mm}
\noindent
\textbf{Pedestrian dataset:} This is built from the  LASIESTA dataset\cite{cuevas2016labeled}, which contains several video sequences taken from a fixed camera with moving persons in the scene. LASIESTA provides ground--truth pedestrian segmentation for each frame in the videos. In this case, ground--truth background  (which is occluded by a person at a given frame) can be found from neighboring frames, after the person has moved away. The foreground map is set to be equal to the segmentation mask provided with the dataset. 
We randomly selected 15 out of 17 video sequences for training, leaving the rest for testing. A sample of 1821 training images was augmented to 45K images via random cropping and color jittering. The test data set contains  198 images.

\subsection{Implementation Details}
Our system was implemented using Tensorflow and  runs on a desktop (3.3Ghz 6-core CPU, 32G RAM, Titan XP GPU). The model was trained with input images resized to $384\times 128$ (for the text dataset images) or  $128\times 128$ (for the pedestrian dataset images.) Adam solver was used during training, with learning rate set to 0.0001,  momentum terms set to $\beta_1=0.5$ and $\beta_2=0.999$, and  batch size equal to 8. We set  $\lambda_f=\lambda_s=100$ in the generator loss (\ref{loss::G}), and followed the standard GAN training strategy \cite{goodfellow2014generative}. Training is alternated between discriminator ($D$) and generator ($G$). Note that the adversarial term for the cost $L_G$ in (\ref{loss::G}) was changed to $(-\log(D(x,y_p)))$ (rather than $\log(1-D(x,y_p))$ ) for better numerical stability, as suggested by Goodfellow \etal \cite{goodfellow2014generative}. When training $D$, the learning rate was divided by 2.

\begin{table}
\begin{center}
\begin{tabular}{c|c|c|c||c|c|c|c}
\hline
Method & \multicolumn{3}{|c||}{Text} & \multicolumn{3}{|c|}{Pedestrian} & Time (s) \\
\cline{2-7}
& $L_1$ & PSNR & SSIM & $L_1$ & PSNR & SSIM &  \\
\hline
\hline
Exemplar ($z_p$) & 5.44\% & 16.563 & 0.553 & 4.11\% & 17.99 & 0.74 & 15.6 \\
Exemplar ($z_g$) & 5.46\% & 16.769 & 0.554 & 3.87\% & 18.36 & 0.77 & \\
\hline
Contextual ($z_p$) & 3.59\% & 19.788 & 0.752 & 3.41\% & 20.916 & 0.879 & 0.23 \\
Contextual ($z_g$) & 3.28\% & 20.170 & 0.779 & 2.64\% & 21.997 & 0.891 & \\
\hline
EPLL ($z_p$) & 2.00\% & 19.123 & 0.732 & 2.9\% & 16.143 & 0.780 & 53.5 \\
EPLL ($z_g$) & 1.87\% & 24.417 & 0.823 & 2.61\% & 16.852 & 0.800 & \\
\hline
IRCNN ($z_p$) & 2.62\% & 21.282 & 0.773 & 2.87\% & 19.117 & 0.870 & 6.67 \\
IRCNN ($z_g$) & {\bf 1.79}\% & {\bf 25.767} & 0.835 & {\bf 2.39}\% & 20.225 & 0.884 & \\
\hline
\hline
Baseline & 2.07\% & 24.188 & 0.811 & 3.06\% & 23.064 & 0.883 & 0.011 \\
\hline
Ours (deconv) & 1.91\% & 24.879 & 0.831 & 2.63\% & 23.612 & 0.904 & 0.018 \\
\hline
Ours (nn+conv) & {1.85\%} & {25.300} & \textbf{0.845} & {2.55\%} & \textbf{23.877} & \textbf{0.918} & 0.018 \\
\hline

\end{tabular}
\end{center}
\caption{Quantitative comparison of our method against other state-of-the-art image inpainting algorithms (Exemplar \cite{criminisi2004region}, Contextual \cite{yu2018generative}, EPLL\cite{zoran2011learning}, and IRCNN\cite{zhang2017learning}). Competing inpainting algorithms are fed with a segmentation mask, either  predicted by our algorithm ($z_p$), or ground--truth ($z_g$). The difference between the original and reconstructed background image is measured using  $L_1$ distance, PSNR (in dB, higher is better) and SSIM (higher is better)\cite{wang2004image}. Time measurements refer to a $128\times 128$ input image.}
\label{tab:comparison}
\end{table}

\subsection{Ablation Study}\label{sec:ablation}
\begin{figure}[h]
\begin{minipage}[b]{.24\linewidth}
  \centering
  \centerline{\epsfig{figure=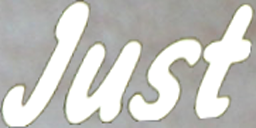,width=3.1cm}}
  \centerline{Input}
\end{minipage}
\hfill
\begin{minipage}[b]{.24\linewidth}
  \centering
  \centerline{\epsfig{figure=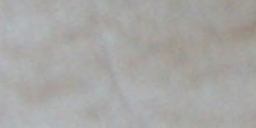,width=3.1cm}}
  \centerline{Ground-truth}
\end{minipage}
\hfill
\begin{minipage}[b]{.24\linewidth}
  \centering
  \centerline{\epsfig{figure=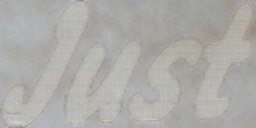,width=3.1cm}}
  \centerline{Baseline}
\end{minipage}
\hfill
\begin{minipage}[b]{.24\linewidth}
  \centering
  \centerline{\epsfig{figure=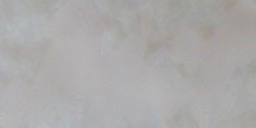,width=3.1cm}}
  \centerline{Ours}
\end{minipage}
\begin{minipage}[b]{.24\linewidth}
  \centering
  \centerline{\epsfig{figure=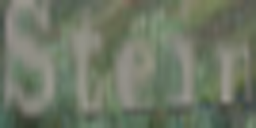,width=3.1cm}}
  \centerline{Input}
\end{minipage}
\hfill
\begin{minipage}[b]{.24\linewidth}
  \centering
  \centerline{\epsfig{figure=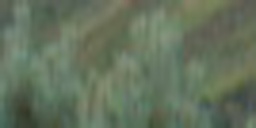,width=3.1cm}}
  \centerline{Ground-truth}
\end{minipage}
\hfill
\begin{minipage}[b]{.24\linewidth}
  \centering
  \centerline{\epsfig{figure=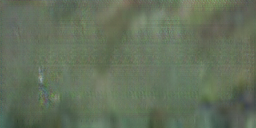,width=3.1cm}}
  \centerline{Ours (deconv)}
\end{minipage}
\hfill
\begin{minipage}[b]{.24\linewidth}
  \centering
  \centerline{\epsfig{figure=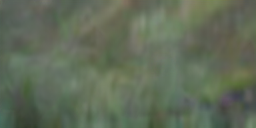,width=3.1cm}}
  \centerline{Ours}
\end{minipage}
\caption{Experimental comparison  between the baseline and our architecture (top), and between our architecture using deconvolution and using upsampling+convolution in {\em dec-fill} (bottom). See Sec. \ref{sec:ablation}.}
\label{fig:ablation}
\end{figure}

\noindent \textbf{Baseline:} In order to validate the effectiveness of the two-branches decoder architecture  and of the neglect layers, we compared our result against a simple baseline structure. This baseline structure is made by the encoder and the {\em dec-fill} decoding branch, without input from the neglect nodes, but with skip layers from the encoder. This is very  very similar to the architecture proposed by Isola \etal \cite{isola2017image}. Tab. \ref{tab:comparison} shows that our method consistently outperforms the baseline structure with both datasets and under  all three evaluation metrics considered ($L_1$ residual, PSNR, SSIM~\cite{wang2004image}). This shows that explicit estimation of the segmentation mask, along with bypass  input from the encoder modulated by the neglect mask, facilitates realistic background image synthesis. An example comparing the result of text removal and of inpainting using the full system and the baseline is shown in the first row of Fig. \ref{fig:ablation}.

\vspace{2mm}
\noindent \textbf{Deconvolution vs. upsampling + convolution:} Deconvolution  (or transpose convolution) is a standard approach for
generating higher resolution images from coarse level features \cite{radford2015unsupervised,donahue2016adversarial,isola2017image}.
A problem with this technique is that it may produce visible checkerboard artifacts, which are due to ``uneven overlapping'' during the deconvolution process, especially when the kernel size is not divisible by the stride. 
Researchers \cite{odena2016deconvolution} have found that by replacing deconvolution  with nearest neighbor upsampling followed by convolution, these artifacts can be significantly reduced. In our experiments, we compared the results using these two techniques (see Fig. \ref{fig:ablation}, second row). Specifically,  upsampling + convolution was implemented using a kernel sized $3\times 3$ with stride $1$ (as described earlier in Sec.~\ref{sec:NA}), while deconvolution was implemented by a kernel with size of $4 \times 4$ and stride of 2. Even though the deconvolution kernel side is divisible by the stride, checkerboard artifacts are still visible in most cases using deconvolution. These artifacts do not appear using upsampling + convolution, which also achieves better quantitative results as shown in Tab. \ref{tab:comparison}.


\subsection{Comparative Results}
Due to the lack of directly comparable methods for automatic content removal, we contrasted our technique with other state-of-the-art image inpainting algorithms, which were provided with a foreground mask. More specifically, we considered two setting for the foreground mask fed to these algorithms: (1) the segmentation mask obtained as a byproduct from our algorithm ($z_p$), and (2) the ground--truth mask ($z_g$). Note that the latter is a best-case scenario for the competing algorithms: our system never accessed this mask. In both cases, the masks were slightly dilated to ensure that the whole foreground region was covered.

Tab. \ref{tab:comparison} shows comparative results with two legacy (but still widely used) inpainting techniques (Exemplar \cite{criminisi2004region} and EPLL\cite{zoran2011learning}), as well as with two more recent CNN-based algorithms (IRCNN\cite{zhang2017learning} and Contextual \cite{yu2018generative}). When fed with the $z_p$ mask (setting (1)), all competing algorithms produced substantially inferior results with respect to ours under all metrics considered. Even when fed with the (unobservable) ground--truth mask $z_g$ (setting (2)), these algorithms generally performed worse than our system (except for IRCNN, which gave better results than ours, under some of the metrics). We should stress that, unlike the competing techniques, our system {\em does not} receive an externally produced foreground map. Note also that our algorithm is faster (often by several orders of magnitude) than the competing techniques.



\begin{figure*}[t]
\begin{center}
\includegraphics[width=\linewidth]{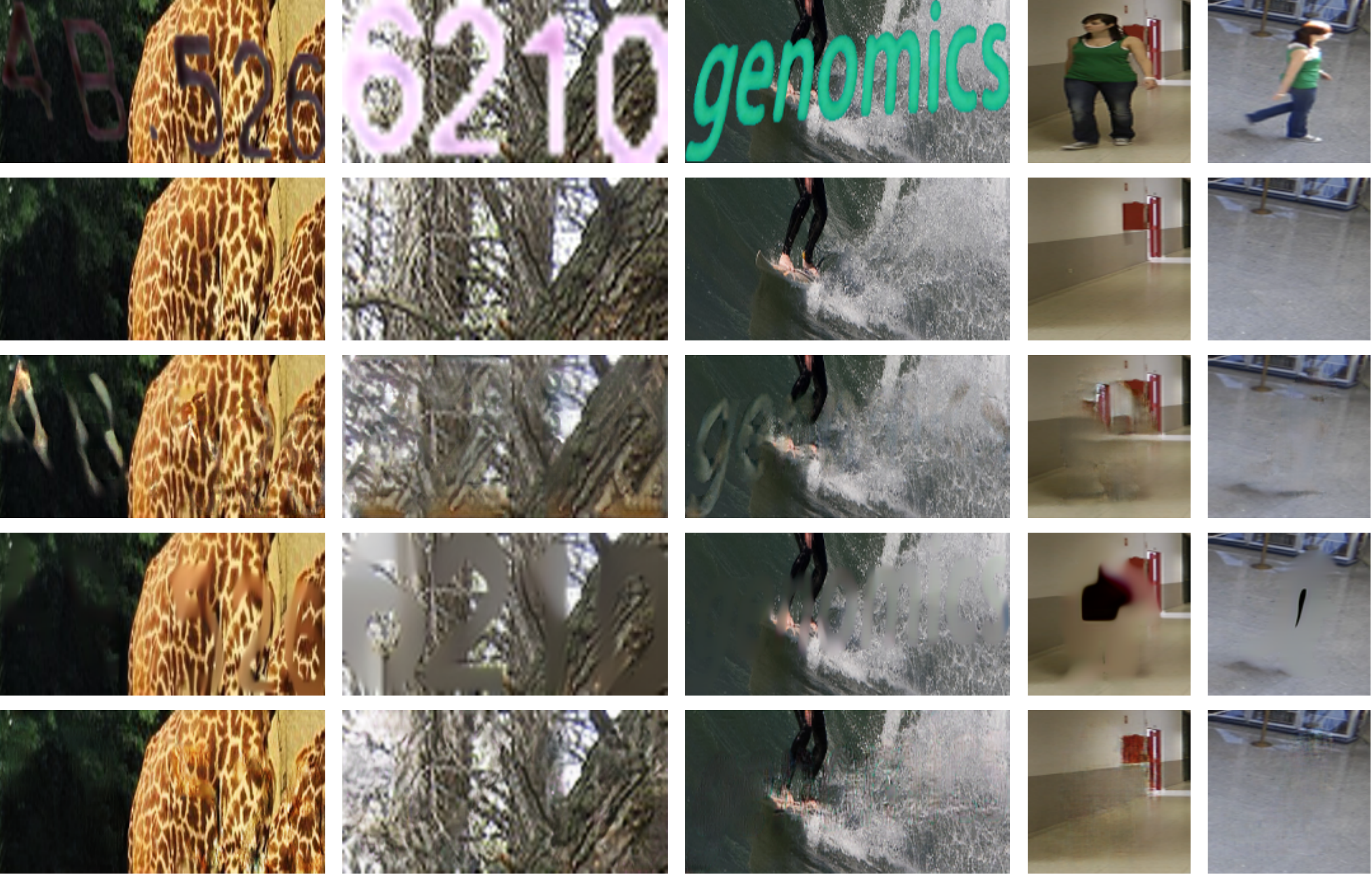}
\end{center}
\caption{Sample inpainting results using Contextual \cite{yu2018generative} (third row), IRCNN \cite{zhang2017learning} (fourth row), and our system (last row). Contextual and IRCNN were fed with the ground--truth segmentation mask, while our system automatically extracted and inpainted the foreground. Top row: input image. Second row: ground-truth background image.}
\label{fig:comp}
\end{figure*}

Fig. \ref{fig:comp} shows comparative examples of results using our system, IRCNN, and Contextual (where the last two were fed with the ground--truth foreground mask, $z_g$). Note that, even when provided with the ``ideal'' mask, the visual quality of the results using these competing methods is generally inferior to that obtained with our content removal technique. 
The result of IRCNN, which is very similar to the result of EPLL, is clearly oversmoothed. This makes the object boundary visible due to the lack of high frequency details in the filled-in region. We also noted that this algorithms cannot cope well with large foreground masks, as can be seen in the last two columns of Fig. \ref{fig:comp} (pedestrian dataset).
 Contextual \cite{yu2018generative}  does a better job at recovering texture,  thanks to its ability to explicitly utilize surrounding image features in its generative model. Yet, we found that our method is often better at completely removing foreground objects. Part of the foreground's boundary  is still visible in  Contextual's reconstructed background region. Furthermore, the quality Contextual's reconstruction drops significantly when the foreground region reaches the border of the image. This problem is not observed with our method.

Fig. \ref{fig:comp} also reveals an interesting (and unexpected) feature of our system. As can be noted in the last two columns,  the shadow cast by the person was removed along with the image of the person. Note that the system was {\em not} trained to detect shadows: the foreground mask only outlined the contour of the person. The most likely reason why the algorithm removed the shadow region is that the  background images in the training set data (which, as mentioned in Sec.~\ref{sec:dataset}, were obtained from frames that did not contain the person) did not contain cast shadows of this type. The system thus decided to synthesize a shadowless image, doubling up as a shadow remover. 
 


\section{Conclusion}
We have presented the first automatic content removal and impainting system that can work with widely different types and sizes of the foreground to be removed and infilled.  Comparison with other state-of-the-art inpainting algorithms (which, unlike are system, need an externally provided foreground mask), along with the ablation study, show that our strategy of joint segmentation and inpainting provides superior results in most cases, at a lower computational cost. Future work will extend this technique to more complex scenarios such as wider ranges of foreground region sizes and transparent foreground.

\subsection*{Acknowledgements}
Research reported in this publication was supported by the National Eye Institute of the National Institutes of Health under award number R01EY029033. The content is solely the responsibility of the authors and does not necessarily represent the official views of the National Institutes of Health.

\bibliography{Siyang_full}
\end{document}